%% file: 0_main.tex
\documentclass[10pt,twocolumn,letterpaper]{article}

\usepackage{iccv}
\usepackage{times}
\usepackage{epsfig}
\usepackage{graphicx}
\usepackage{amsmath}
\usepackage{amssymb}
\usepackage[ruled]{algorithm2e}

\usepackage{mathrsfs}
\usepackage{multirow}

\usepackage[pagebackref=true,breaklinks=true,letterpaper=true,colorlinks,bookmarks=false]{hyperref}

\iccvfinalcopy 

\def\iccvPaperID{832} 
\vspace{-20pt}
\ificcvfinal\fi

\begin{document}

\title{Permutation-invariant Feature Restructuring for Correlation-aware \\Image Set-based Recognition}

\author{Xiaofeng Liu{$^{1,2}$}, Zhenhua Guo{$^{1,4}$}, Site Li{$^{1}$}, Lingsheng Kong{$^{3}$}, Ping Jia{$^{3}$}, Jane You{$^{5}$}, B.V.K. Kumar{$^{1*}$}\\
{$^{1}$}Carnegie Mellon University;~ {$^{2}$}Harvard University;~ {$^{3}$}CIOMP, Chinese Academy of Sciences;\\ {$^{4}$}Graduate School at Shenzhen, Tsinghua University;~ {$^{5}$}The Hong Kong Polytechnic University\\
{$^{*}$}Corresponding author: kumar@ece.cmu.edu
}

\maketitle

\input{1_Abstract.tex}

\input{2_Introduction.tex}

\input{3_RelatedWork.tex}

\input{4_Approach1.tex}

\input{4_Approach2.tex}

\input{4_Approach3.tex}

\input{4_Approach4.tex}

\input{5_Experiments1.tex}

\input{5_Experiments2.tex}

\input{5_Experiments3.tex}

\input{5_Experiments4.tex}

\input{6_Conclusions.tex}

{\small
\bibliographystyle{ieee}
\bibliography{egbib}
}

\end{document}

%% file: 1_Abstract.tex
\begin{abstract}
We consider the problem of comparing the similarity of image sets with variable-quantity, quality and un-ordered heterogeneous images. We use feature restructuring to exploit the correlations of both inner$\&$inter-set images. Specifically, the residual self-attention can effectively restructure the features using the other features within a set to emphasize the discriminative images and eliminate the redundancy. Then, a sparse/collaborative learning-based dependency-guided representation scheme reconstructs the probe features conditional to the gallery features in order to adaptively align the two sets. This enables our framework to be compatible with both verification and open-set identification. We show that the parametric self-attention network and non-parametric dictionary learning can be trained end-to-end by a unified alternative optimization scheme, and that the full framework is permutation-invariant. In the numerical experiments we conducted, our method achieves top performance on competitive image set/video-based face recognition and person re-identification benchmarks.
\end{abstract}

%% file: 2_Introduction.tex
\section{Introduction}
\begin{figure}\setlength{\abovedisplayskip}{-5mm}\setlength{\belowdisplayskip}{-10mm}
\centering
\includegraphics[width=8cm]{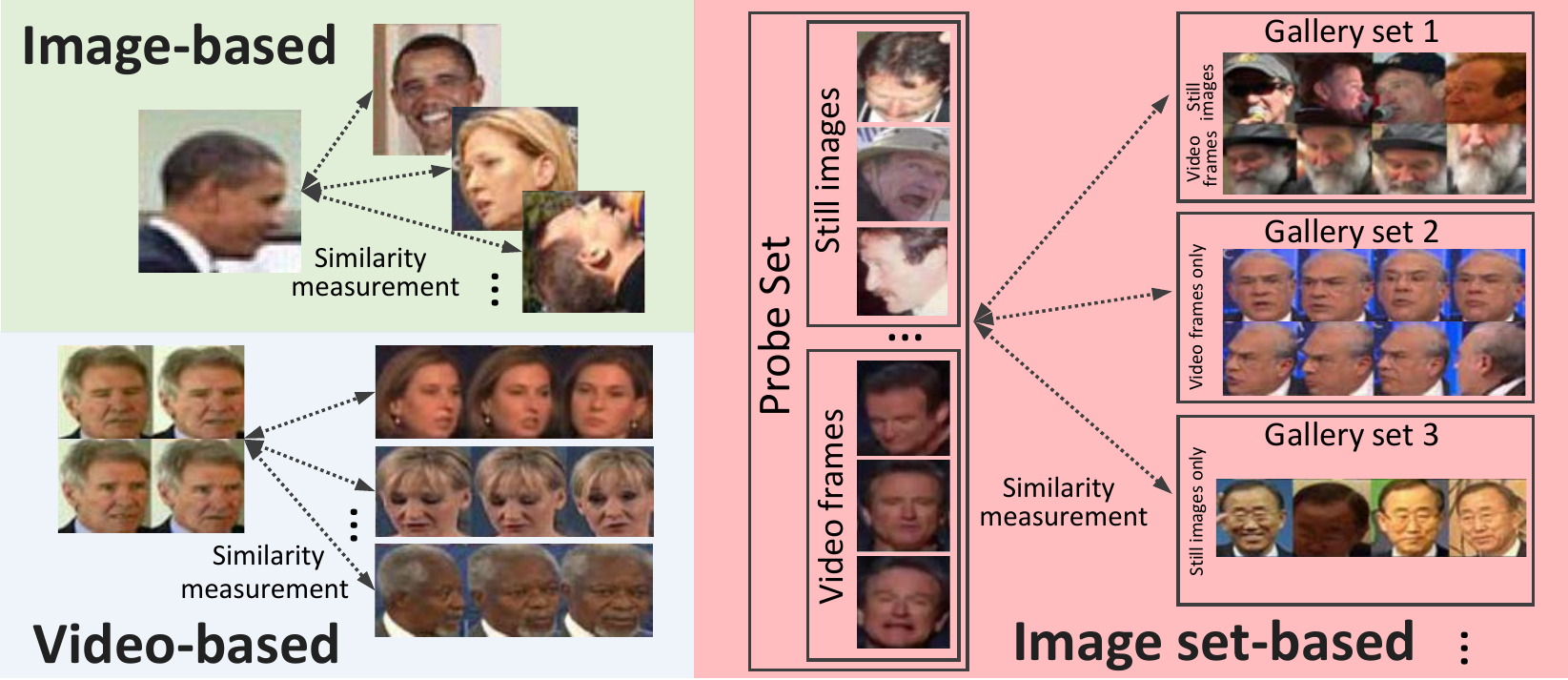}\\
\caption{Illustration of the single image-based, video sequence-based and image set-based open-set face identification.}
\label{fig:1}
\end{figure}

Many research studies focus on using a single image \cite{liu2019feature,liu2017adaptive,liu2019hard} or video \cite{liu2018adaptive,liu2017line} as shown in Fig. \ref{fig:1}. However, in many practical applications, a set of images of a subject (consisting of still images, or frames from a video, or a mixture of both) can usually be collected from different checkpoints, segments in videos, mugshot history of a criminal and lifetime enrollment images for identity documents $etc$ \cite{liu2019dependency}. These can contain extreme rotations, complex expressions and illumination variations \cite{liu2018dependency,liu2017quality,yang2017neural}. This setting is more similar to the real-world biometric scenarios \cite{grother2014face}.

Compared to the case of a single image \cite{liu2019research,liu2018data,liu2018normalized,liu2018ordinal,liu2019unimodala,liu2019unimodalb}, richer and complementary information can be expected in a set, because the samples are captured from multiple views \cite{liu2019conservative}. However, it also poses several challenges including a) $variable$ $number$ of samples within a set, b) larger $inner$-$set$ $variability$ than its video-based recognition counterpart, and c) $order$-$less$ $data$.

\begin{figure*}[t!]
\centering
\includegraphics[width=17.5cm]{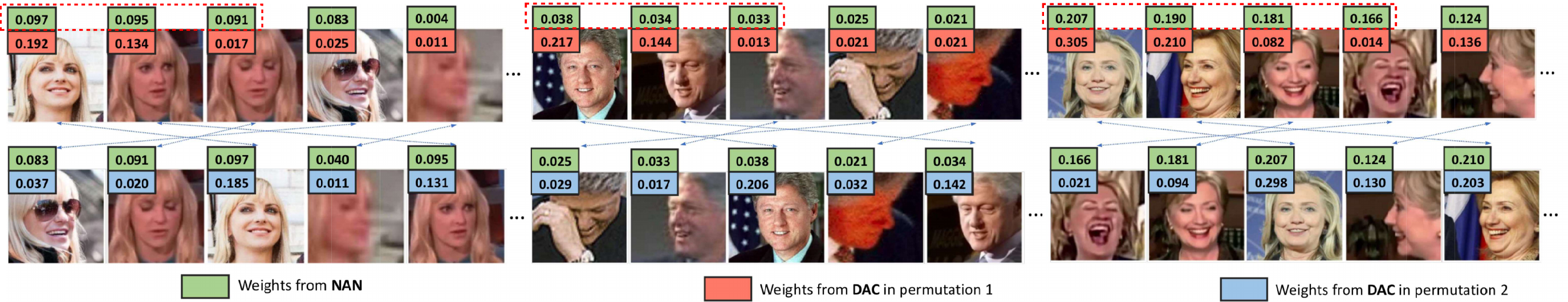}\\
\caption{Examples in IJB-A dataset showing the weights of images calculated by independent image quality assessment (NAN \cite{yang2017neural}) that does not take into account inner-set correlations and RL-based inner-set relationship exploring (DAC \cite{liu2018dependency}). The NAN suggests a similar score for a discriminative image and its blurry version in the same pose (dashed red box). However, there is little additional information in the latter. When we change the order of images from the 1st row to the 2nd row, the output of DAC is different. If the inferior samples are processed ahead by DAC, their weights can be larger, and the performance would be degraded.} 
\label{fig:2}
\end{figure*}

The conventional approaches to set-based recognition aggregate all images X$={\left\{x_n\right\}}_{n=1}^N$ in a set with $N$ images to a single feature vector. The aggregation is achieved through max/average pooling \cite{li2014eigen,parkhi2015deep,chen2015end}. More formally, $f({\rm X})=\rho(max/average~pooling(\left\{\varphi(x_1),\cdots,\varphi(x_N)\right\})$, where $\rho, \varphi$ are any appropriate mapping functions. Considering that the importance of each image maybe different, their weights are calculated independently by $\varphi$ as the soft attention score in pooling operation \cite{yang2017neural,liu2017quality}. By doing this, the high quality ($i.e.,$ clear and frontal) faces are favored in these models. These methods are not sensitive to the order of images with in a set (i.e., permutation-invariant), but their weight selection procedures for each image in the set do not pay attention to the other images. Without inner-set correlation can result in redundant weights as in Fig. \ref{fig:2}. 

The inter-set correlation is also important for similarity measurements \cite{yin2017multi}. If we are given a query set consisting of clear profile face images, it should be more reasonable to emphasize the profile face images in the gallery set rather than the high-quality frontal face images.

How to model the inner/inter-set correlations in image sets with arbitrary number of order-less images has long been a challenging \cite{liu2018dependency}.

For exploring the inner-set correlations, the typical permutation-sensitive methods, $e.g.,$ recurrent neural networks, are totally not applicable for order-less data \cite{liu2018dependency}. The reinforcement learning (RL) was introduced to estimate the weights of each image sequentially \cite{liu2018dependency}. Fig. \ref{fig:2} shows that RL is still sensitive to the order.

For exploring the inter-set correlations, the parametric pair-wise input network (takes two sets as input and output the similarity score using a sigmoid unit directly) usually computationally inefficient for open-set identification \cite{xie2018comparator,yang2017neural}. \cite{liu2018dependency} proposes to represent and compare frontal and profile faces separately. When we start using separate representations for expressions, lighting, occlusion, makeup $etc$., the enormous number of final representations and hyper-parameters needed to balance them will be uncontrollable.

To address the challenge of taking inner-set and inter-set correlations into account in a computationally efficient way, we propose the \textit{permutation-invariant feature restructuring} (\textbf{PIFR}) framework to reduce the redundancy and maintain the benefit of diversity information. It fully considers the inner/inter-set correlation by restructuring a sample in the probe set with the samples in probe/gallery set to explore their complementarity and substitutability.

Specifically, a {parametric redundancy-eliminating self-attention} ({RSA}) module allows all samples in a set to contribute to produce a refined feature vector based on feature pair-wise affinity. Its residual attention mechanism not only learns to emphasize discriminative samples while repelling inferior ones, but it also explicitly reduces the redundancy. Moreover, spatial clues are incorporated using the {fully convolutional feature extractor} and Gaussian similarity metric.

To take all possible inter-set variations into account and scale to open-set identification, we develop a {non-parametric dependency-guided feature alignment} ({DFA}) module for inter-set interaction. It assumes that each probe image feature vector can be sparsely/collaboratively restructured on the basis of gallery feature set \cite{liu2018joint}. In testing, the gallery features can be pre-computed and stored, and its reconstruction coefficients can be computed in parallel.

The parametric deep learning-based RSA and non-parametric dictionary learning-based DFA can be alternatively optimized in our PIFR. Table \ref{tab:1} compares the capabilities of the proposed PIFR to other set-based approaches. The main contributions of this paper are

\begin{figure*}[t]
\centering
\includegraphics[width=17cm]{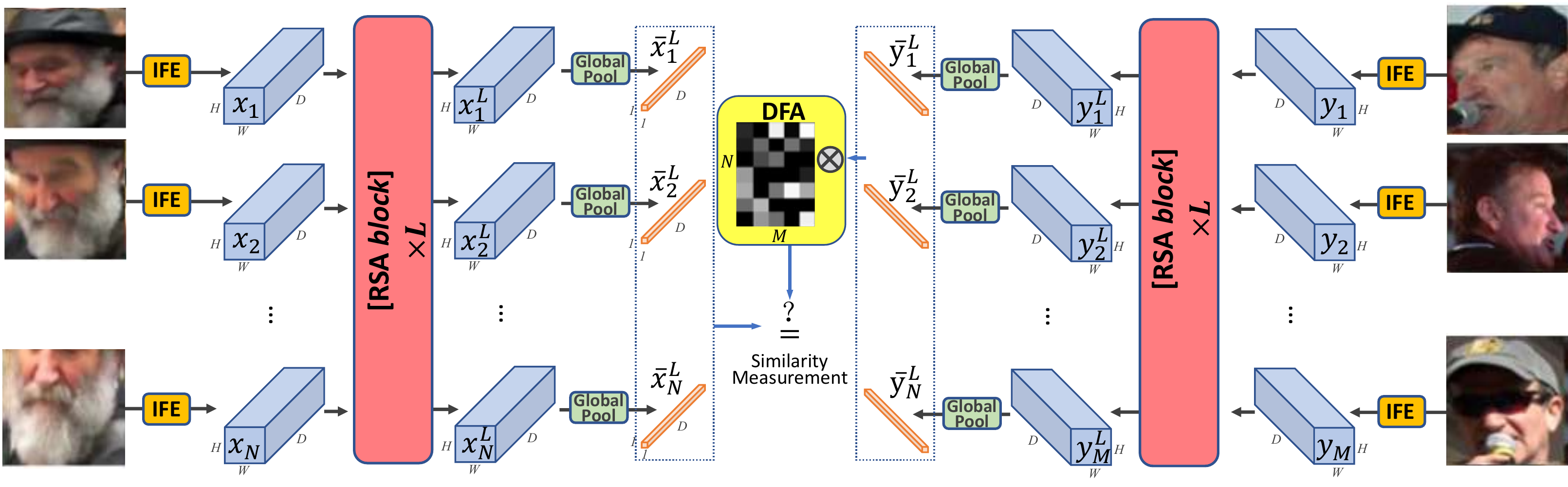}\\
\caption{Our \textit{permutation-invariant feature restructuring} (\textbf{PIFR}) framework for image set-based recognition, which consists of the fully convolutional image feature extractor (IFE), redundancy-eliminating self-attention blocks for inner-set restructuring (RSA$\times L$), global pooling layer and sparse/collaborative representation-based dependency-guided feature alignment (DFA) module for inter-set restructuring.}
\label{fig:3}
\end{figure*}\begin{table}
\begin{center}
\tiny
\begin{tabular}{|l|c|c|c|c|c|c|}
\hline
\multirow{2}*{Method} & permutation &inner-set &inter-set &\multirow{2}*{verif-} &open-set \\
& invariant &correlation&correlation&ication&identification\\\cline{2-5}
\hline\hline
AvePool&yes& & &yes&yes\\\hline
NAN\cite{yang2017neural} & yes& & &yes&yes\\\hline
DAC\cite{liu2018dependency} & &yes&pose-only&yes&yes\\\hline
CNet\cite{xie2018comparator} &yes&pose-only&pose-only&yes&~\\\hline\hline
\textbf{PIFR}&\textbf{yes}&\textbf{yes}&\textbf{yes}&\textbf{yes}&\textbf{yes}\\\hline
\end{tabular}
\end{center}
\vspace{-7pt}
\caption{Comparisons between proposed PIFR and other state-of-the-art approaches for set-based recognition.}\label{tab:1}
\end{table}

$\bullet$ We introduce PIFR that takes into account the correlations among variable number of order-less samples by restructuring the feature with the other inner$\&$inter-set features in a permutation-invariant manner. It emphasizes the discriminative images, reduces the redundancy, adaptively aligns the probe-gallery set, and is efficient for both verification and open-set identification tasks.

$\bullet$ We show how self-attention and dictionary learning can be applied to set-based recognition. The spatial cues and explicit redundancy-reduction are further incorporated in RSA. A novel reconstruction-based sparse/collaborative similarity and loss are proposed in DFA.

$\bullet$ The proposed PIFR with alternate optimization can be a general framework to integrate the parametric deep learning and non-parametric dictionary learning.

We apply the proposed method to several image set-based verification/identification tasks and empirically demonstrate its effectiveness and generality.

%% file: 3_RelatedWork.tex
\section{Related Works}

\noindent\textbf{Representation of sets.} The most straightforward permuta-tion-invariant operations are the max/average pooling over a set of embeddings X=${\left\{x_n\right\}}_{n=1}^N$ \cite{li2014eigen,parkhi2015deep,chen2015end}. Zaheer $et~al.$ \cite{zaheer2017deep} has proven that the function that can be represented the following form is permutation-invariant:\begin{equation}  
f({\rm X})=\rho(\sum_{x\in {\rm X}} (\varphi(x))\label{con:1} 
\end{equation} However, $\varphi$ acts independently on each
element, which does not allow an exploration of the inter-sample relationships. This limitation is also in the methods in \cite{edwards2016towards,cotter2018interpretable}.

\noindent \textbf{Image set based recognition.} A video can be considered as an image set with ordered images and has been actively studied, while a general image set is orderless and usually contains more challenging variations. Thus the methods relying on temporal dynamics will not be covered here. Traditional methods usually model the set/video as appearance subspaces or manifolds \cite{wang2015discriminant,huang2016building,wang2012covariance}.

Along a different axis, \cite{yang2017neural,liu2017quality} are based on computing a score for each image with neural image assessment modules. Then, a set of features are aggregated to a fixed size feature vector via weighted pooling. Without inner-set interactions, this may result in redundancy and not use the information diversity present in a set.

\cite{liu2018dependency} proposes to exploit the inner-set relationship using the RL. However, it is permutation-variant since the weight of the first image is computed based on $[x_1;\frac{x_2+\cdots +x_N}{N-1}]$, while if we move that image to the second position, its weight will be based on the $[x_2;\frac{a_1x_1+x_3+\cdots +x_N}{N-2+a_1}]$, where $[\cdot;\cdot]$ denotes the concatenation and $a_1$ is the updated weight of the first image. Besides, the RL itself is usually unstable \cite{henderson2017deep}. The feature dimension is compressed from 1024 in its GoogleNet backbone \cite{szegedy2016rethinking} to 128 which unavoidably weakens its representation ability. Different from these extensions, we address the inner-set redundancy by constructing a feature vector that takes into account all of its neighbors and results in a set of redundancy-reduced features. 

Using a single set-level feature vector is inadequate for representing inter-set interactions. \cite{liu2018dependency,xie2018comparator} target the inter-set pose-aware alignment relying on pose/landmark detection, and the parametric pair-wise input network in \cite{xie2018comparator} is computationally inefficient for open-set identification. Our PIFR is end-to-end trained given only the set-level identity annotations, and is compatible with multiple inter-set variations and is scalable for open-set identification.

\noindent \textbf{Self-attention and non-local filtering.}
As attention models grow in popularity, \cite{vaswani2017attention} develops a self-attention mechanism for machine translation. It calculates the response at one position as a weighted sum of all positions in the sentences. A similar idea is also inherited in the non-local algorithm \cite{buades2005non}, which is a classical image denoising technique. The interaction networks were also developed for modeling pair-wise interactions \cite{battaglia2016interaction,hoshen2017vain,watters2017visual,yang2018learning}. Moreover, \cite{wang2018non} proposes to bridge self-attention to the more general non-local filtering operations. \cite{zhou2017temporal} proposes to learn temporal dependencies between video frames at multiple time scales. Inspired by above works, we further adapt this idea to the set-based problem, the difference with the aforementioned studies includes, but is not limited to incorporating spatial clues with Gaussian similarity matrix, explicit redundancy-eliminating residual term and single-layer pairwise affinity.

\noindent \textbf{Sparse/collaborative representation based classification (SRC/CRC).} SRC \cite{wright2009robust} and CRC \cite{zhang2011sparse}, and their many extensions \cite{liu2018joint,chi2014classification,wu2014collaborative} have been widely studied. They achieve promising single image and video-based \cite{chen2012dictionary} face recognition performance under occlusions and illumination variations. Our DFA is inspired by them, but using a novel similarity measurement instead of reconstruction coefficients.

%% file: 4_Approach1.tex
\section{Methodology}

We consider the image-set based recognition as matching the probe set P containing $N$ images ($i.e.,$ P$={\left\{p_n\right\}}_{n=1}^N$) with the gallery set G which consist of $M$ images ($i.e.,$ G $={\left\{g_m\right\}}_{m=1}^M$), $n,m$ are the image/video frames indexes. As illustrated in Fig. \ref{fig:3}, our permutation-invariant feature restructuring (PIFR) framework consists of three major components: fully convolutional image feature extractor (IFE), a group of redundancy-eliminating self attention (RSA) blocks and a dependency-guided feature alignment (DFA) module. To represent the spatial correlations, we choose the convolution layers of a deep convolutional neural network (CNN) networks with high-end recognition performances ($e.g.$, GoogleNet\cite{szegedy2016rethinking}, ResNet50\cite{he2016deep}) as our IFE, instead of either hand-crafted extractors or CNN with fully connected layers used in \cite{liu2018dependency,yang2017neural}. It embeds the images into latent space independently and generates the corresponding feature sets X$={\left\{x_n\right\}}_{n=1}^N$ and Y$={\left\{y_m\right\}}_{m=1}^M$ from P and G respectively. $x_n,y_m \in \mathbb{R}^{H\times W\times D}$ where $H$, $W$ and $D$ are the height, width and channel dimension, respectively of our representation. Its parameters are fixed after pre-training by a large scale image-based recognition datasets.

\begin{table}
\tiny
\newcommand{\tabincell}[2]{\begin{tabular}{@{}#1@{}}#2\end{tabular}}
\begin{center}
\begin{tabular}{|c|c|c|c|}
\cline{2-4}

\multicolumn{1}{c|}{~}& \multicolumn{2}{|c|}{Layers} & Output size\\
\cline{2-4} \hline
 \multirow{5}*{\rotatebox{90}{Image Feature Extractor~~~~~~~}}& $conv$&$7\times7,64$, stride 2& $112\times112\times64\times$$N$\\
\cline{2-4}
 & $pool$& max pool, 3$\times$3, stride 2&$56\times56\times64\times$$N$\\
\cline{2-4}
 &\tabincell{c}{1-3 $res$\\$blocks$}&
 $\begin{bmatrix}
   {\rm conv}&1\times1,64 \\
   {\rm conv}&3\times3,64 \\
   {\rm conv}&1\times1,256
  \end{bmatrix}$$\times$3 &$56\times56\times256\times$$N$\\
\cline{2-4}
 & \tabincell{c}{4-7 $res$\\$blocks$}& $\begin{bmatrix}
   {\rm conv}&1\times1,128 \\
   {\rm conv}&3\times3,128 \\
   {\rm conv}&1\times1,512
  \end{bmatrix}$$\times$4&$28\times28\times512\times$$N$\\
\cline{2-4}
 & \tabincell{c}{8-13 $res$\\$blocks$}& $\begin{bmatrix}
   {\rm conv}&1\times1,256 \\
   {\rm conv}&3\times3,256 \\
   {\rm conv}&1\times1,1024
  \end{bmatrix}$$\times$6&
  \tabincell{c}{$14\times14\times1024\times$$N$\\$(H\times W\times D\times N)$} \\
\hline
\multicolumn{2}{|c|}{RSA}&$\begin{bmatrix}
   {\rm RSA block}
  \end{bmatrix}$$\times L$ &$14\times14\times1024\times$$N$\\
\hline
\multicolumn{2}{|c|}{Global pool}&Global pool, 14$\times$14&$1\times1\times1024\times$$N$\\
\hline
\end{tabular}
\end{center}
\caption{The structure details of our IFE and RSA modules. We adopt the pre-trained ResNet-50 backbone, and utilize the output of $3^{rd}$ resblock as our ${\left\{x_n\right\}}_{n=1}^N$ on IJB-A\&Celebrity-1000.}
\end{table}

%% file: 4_Approach2.tex
\subsection{Redundancy-eliminating Self Attention}
\begin{figure}[t]
\centering
\includegraphics[width=8cm]{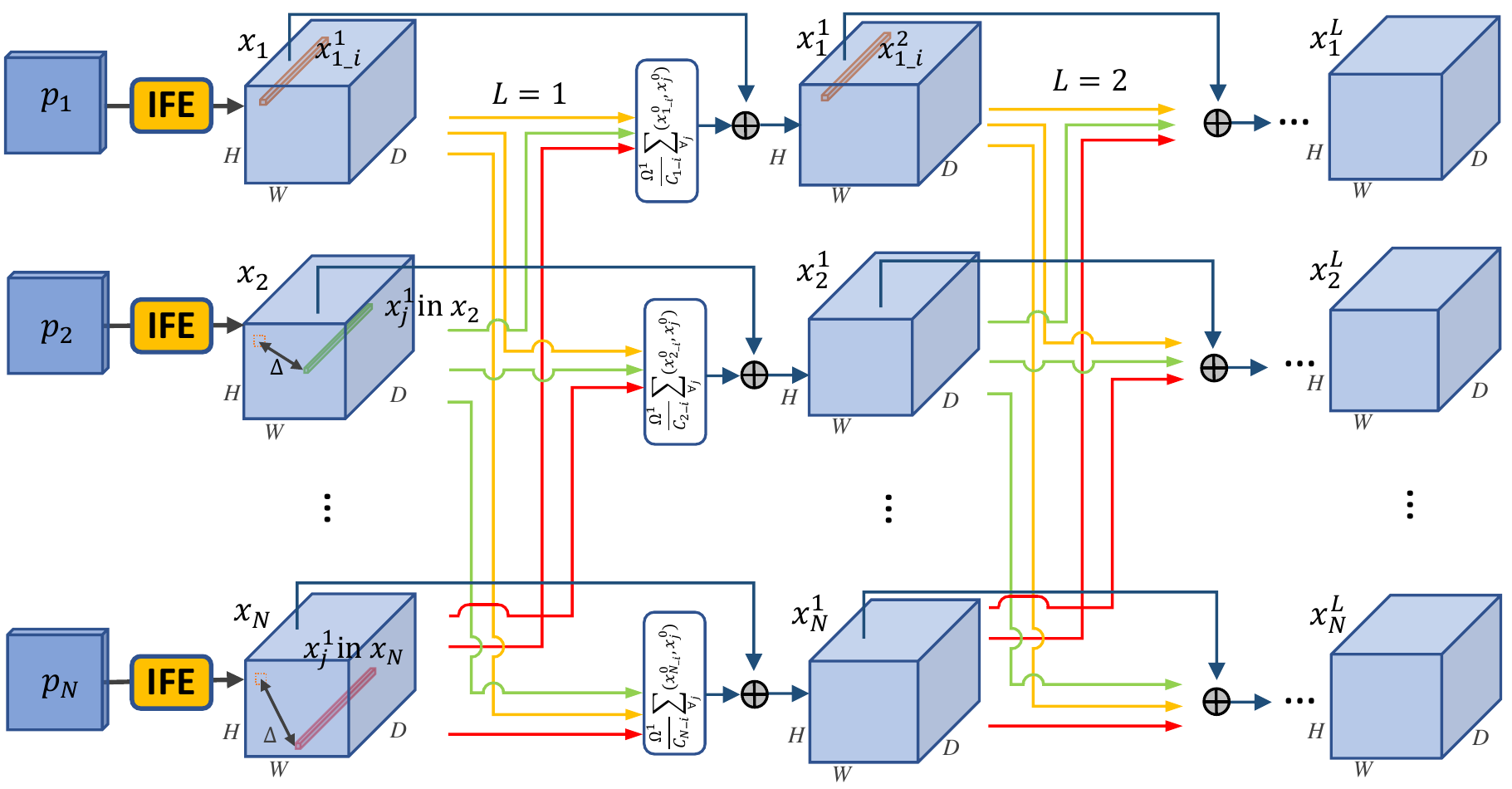}\\
\caption{Redundancy-eliminating self attention blocks.}
\label{fig:4444}
\end{figure}

A shown in Fig. \ref{fig:4444}, we cascade $L\times$ RSA blocks, which works as a residual self-attention module receiving all extracted features ${\left\{x_n\right\}}_{n=1}^N$ and restructuring them based on inner-set correlations. Since the features ${\left\{x_n\right\}}_{n=1}^N$ are deterministically computed from the probe set ${\left\{p_n\right\}}_{n=1}^N$, they also inherit and display large variations and redundancy. Ideally, the RSA should display four attributes: Ability to handle arbitrary number of samples, inner-set correlation-awareness, permutation invariance to the features, and sensitivity to  relative locations in the image plane.

We propose to harvest the spatially-aware inner-set correlations by exploiting the affinity of point-wise feature vectors. We use $i=1,\cdots,{\small H}\times {\small W}$ to index the position in HW plane and the $j$ is the index for all $D$-dimensional feature vectors other than the $i^{th}$ vector ($j=1,\cdots,{\small H}\times {\small W}\times {\small N}-1)$. Specifically, our RSA block can be formulated as\begin{equation}\vspace{-5pt}x_{n\_i}^{l}=x_{n\_i}^{l-1}+\frac{\Omega^l}{C_{n\_i}}\sum_{\forall {n\_j}}\omega(x_{n\_i}^{0},x_{n\_j}^{0})(x_{n\_j}^{l-1}-x_{n\_i}^{l-1})\Delta_{i,j}\nonumber\label{con:2}\vspace{-10pt}\end{equation}\begin{align}{C_{n\_i}}=\sum_{\forall n\_j}\omega(x_{n\_i}^{0},x_{n\_j}^{0})\Delta_{i,j}\vspace{-5pt}\end{align}where ${\Omega^l}\in \mathbb{R}^{1\times1\times D}$ is the weight vector to be learned, $l=0,1,\cdots,L$ with $L$ being the number of stacked sub-self attention blocks and ${x}_n^0={x_n}$. The pairwise affinity $\omega(\cdot,\cdot)$ is an scalar. The response is normalized by ${C_{n\_i}}$. 

To explore the spatial clues, we propose to use a “Gaussian similarity” metric $\Delta_{i,j}={\rm exp}{(\frac{{\parallel hw_{n\_i}-hw_{n\_j} \parallel}_2^2}{\sigma })}$ , where $hw_{n\_i}$, $hw_{n\_j}\in\mathbb{R}^2$ represent the position of $i^{th}$ and $j^{th}$ vector in the HW-plane of $x_n$, respectively.

The residual term is the difference between the neighboring feature ($i.e.,x_{n\_j}^{l-1}$) and the computed feature $x_{n\_i}^{l-1}$. If $x_{n\_j}^{l-1}$ incorporates complementary information and has better imaging/content quality compared to $x_{n\_i}^{l-1}$, then RSA will erase some information of the inferior $x_{n\_i}^{l-1}$ and replaces it by the more discriminative feature representation $x_{n\_j}^{l-1}$. Compared to the method of using only $x_{n\_j}^{l-1}$ \cite{wang2018non,zhou2017temporal}, our setting shares more common features with diffusion maps \cite{coifman2006diffusion,tao2018nonlocal}, graph Laplacian \cite{chung1997spectral} and non-local image processing \cite{buades2005review,gilboa2007nonlocal}. All of them are non-local analogues \cite{du2012analysis} of local diffusions, which are expected to be more stable than its original non-local counterpart \cite{buades2005non,wang2018non} due to the nature of its inherit Hilbert-Schmidt operator \cite{du2012analysis}.

The operation of $\omega$ in Eq. \eqref{con:2} is not sensitive to many function choices \cite{wang2018non,zhou2017temporal}. We simply choose the embedded Gaussian given by\vspace{-10pt}\begin{align}
\omega(x_{n\_i}^{0},x_{n\_j}^{0})=e^{\psi{(x_{n\_i}^{0})}^T\phi(x_{n\_j}^{0})}
\end{align}\noindent where $\psi({x_{n\_i}^{0})={\Psi}x_{n\_i}^0}$ and $\phi(x_{n\_j}^{0})={\Phi}x_{n\_j}^0$ are two embeddings, and $\Psi$, $\Phi$ are matrices to be learned.

Different from \cite{wang2018non}, our affinity $\omega$ is pre-computed given the input feature set ${\left\{x_n\right\}}_{n=1}^N$ and stays the same through the propagation within RSA, which reduces the computational cost while $\omega(x_{n\_i}^{0},x_{n\_j}^{0})$ can still represent the affinity between $x_{n\_i}^{l-1}$ and $x_{n\_j}^{l-1}$ to some extent. This design reduces the number of parameters and speeds up the learning.

\noindent \textbf{Definition 1.} Let $\pi$ be an arbitrary permutation function for a sequence. We say a function $f$(X) is $permutation$-$invariant$ iff for $\forall\pi$; $f$(X$)=f(\pi$(X)).

\noindent \textbf{Property 1.} The RSA is $permutation$-$invariant$, due to the fact that all features in ${\left\{x_n\right\}}_{n=1}^N$ are considered in {Eq. \eqref{con:2}} in an order-less manner (sum operation).

The formulation in Eq. \eqref{con:2} supports the inputs of variable number of features, and maintains the corresponding number and dimension of features in the output tensor ${\rm \footnotesize {X}}^L$ after several blocks are calculated consecutively. We apply the global pooling (GP) \cite{lin2013network} to summarize the information in HW-plane, and get the final representations:

\begin{equation}  \setlength{\abovedisplayskip}{-3mm}
\left\{  
            \begin{array}{lr}  
            {\rm\overline{X}}^L=\left\{\overline{x}_1^{L},\cdots,\overline{x}_N^{L}\right\},\overline{x}_n^L={\rm GP}({x}_n^L)\in \mathbb{R}^{1\times 1\times D}\\ {\tiny~}\vspace{-5pt}\\  
            {\rm\overline{Y}}^L=\left\{\overline{y}_1^{L},\cdots,\overline{y}_M^{L}\right\},\overline{y}_m^L={\rm GP}({y}_m^L)\in \mathbb{R}^{1\times 1\times D}    
            \end{array} 
\right.
\end{equation}

RSA learns to aggregate inner-set correlations for each individual position of a sample's feature adaptively. It can be flexibly attached to any pre-trained IFE (by initializing $\Omega^l$ as zero) and possible to be fine-tuned jointly with IFE.

%% file: 4_Approach3.tex
\subsection{Dependency-guided Feature Alignment}
Beyond specifically aligning the poses, illuminations, expressions $etc$., \cite{liu2018dependency,yin2017multi}, we propose to measure the dependency-guided similarity between a pair of feature sets with different feature numbers leveraging the non-parametric dictionary learning methods \cite{liu2018joint}. 

$\overline{x}_n^{L}$ is expected to be represented using a linear combination of $\left\{\overline{y}_1^{L},\cdots,\overline{y}_M^{L}\right\}$, which means that we can search similar feature vectors in ${\rm \footnotesize\overline{Y}}^L$ to reconstruct $\overline{x}_n^{L}$. It transforms the matching task into a sparse/collaborative representation problem:\vspace{-6pt}\begin{align} \setlength{\abovedisplayskip}{-5mm}\setlength{\belowdisplayskip}{-5pt}
\mathop{}_{a_n}^{\rm min}{\parallel{a_n}\parallel}_p  ~~~~ s.t.~~~~ \overline{x}_n^{L}= {\rm \footnotesize\overline{Y}}^L{a_n}\label{con:5}
\end{align} where ${a_n}\in \mathbb{R}^{M \times 1}$ is the sparse/collaborative coding vector of $\overline{x}_n^{L}$ $w.r.t.$ ${\rm \footnotesize\overline{Y}}^L$. The ${\ell}_p$-norm is used to constrain ${a_n}$, which leads to fewer features in ${\rm \footnotesize\overline{Y}}^L$ to reconstruct the $\overline{x}_n^{L}$. $p=1$ and $2$ are adopted in sparse and collaborative representation-based classification, respectively. However, SRC/CRC is proposed to reconstruct a probe image based on all of the gallery images and uses the coefficient $a_n$ to rank similar samples. Our method is different from that, since the objectives are the samples in the sets and we do not use ${\parallel{a_n}\parallel}_p$ directly for similarity measurement. We can relax the constraint term and rewrite Eq. \eqref{con:5} as \vspace{-4pt}\begin{align}\setlength{\abovedisplayskip}{-5mm}\setlength{\belowdisplayskip}{-5pt}
\mathop{}_{a_n}^{\rm min}{\Vert \overline{x}_n^{L} - {\rm \footnotesize\overline{Y}}^L{a_n}\parallel}{}_2^2 +\frac{\lambda}{{M}}{\Vert{a_n}\parallel}_p \label{con:6}
\end{align}where ${\parallel \overline{x}_n^{L}-{\rm \footnotesize\overline{Y}}^L{a_n}\parallel}{}_2^2$ is the ${\ell}_2$ distance of $\overline{x}_n^{L}$ and ${\rm \footnotesize\overline{Y}}^L$, and $\lambda$ controls the sparsity of coefficients $a_n$. For a probe set with $N$ samples, we need a sparse/collaborative reconstruction coefficient matrix ${\rm A}=\left\{a_1,\cdots,a_N\right\}\in \mathbb{R}^{M\times N}$, and the set-to-set verification is reformulated to calculate the set-level similarity, which can be measured by $ \frac{1}{N}{\parallel{\footnotesize\rm \overline{X}}^L - {\rm \footnotesize\overline{Y}}^L{\rm A}\parallel}{}_2^2$. It is essentially different from previous metric learning-based algorithms for two sets.

As such, DFA is essentially accounting for all possible variations to further boost the performance at the extremes of these variations, and there is no need for additional pose/landmark annotations. Since there are no additional constraints for the relations of the elements in A, it is also permutation-invariant.

In the testing stage of an open-set identification task with $K$ gallery sets, we need $K$ times verification, where $K$ can be a few hundreds ($e.g.,$ IJB-A) or even millions in real world applications.The pair-wise input network \cite{xie2018comparator} for inter-set interaction need to process $K$ query-gallery image set pairs. Considering the limited memory of GPUs, we usually cannot put in 2$K$ image sets and several network models. Its processing time equals to $K$ times of verification, which is impractical. However, our gallery ${\rm\scriptsize\overline{Y}}^L$s can be pre-computed and stored. We only need to extract ${\rm\scriptsize\overline{X}}^L$ from a query set, and the $K$ times similarity measurements is applied to ${\rm\scriptsize\overline{X}}^L$ and ${\rm\scriptsize\overline{Y}}^L$ vector set pairs instead of original image. Unlike parametric network-based scheme, our DFA  does not take a lot of memory. Therefore, we can process numerous ${\rm\scriptsize\overline{X}}^L$ and ${\rm\scriptsize\overline{Y}}^L$ pairs in parallel. Actually, $a_1,\cdots a_N$ are also solved using a parallel algorithm \cite{lee2007efficient,zhang2011sparse}. The separated and non-parametric DFA enables our framework to be practicable for open-set identification.

\begin{algorithm}[!tp] \setlength{\abovedisplayskip}{-5mm}\setlength{\belowdisplayskip}{-5mm}
\small
\caption{Bi-level training of PIFR}
\LinesNumbered 
\KwIn{Training sets and labels; $\lambda$; Pre-trained IFE}
\KwOut{Updated RSA parameters $\theta$}
\textbf{Level 1: Pre-train RSA}\; 
\For{all {\rm{IFE}} extracted training sets {\rm{X}} and {\rm{Y}}}{
Reconstruct ${\left\{x_n\right\}}_{n=1}^N$, ${\left\{y_m\right\}}_{m=1}^M$ to ${\rm\footnotesize{X}}^L$,${\rm\footnotesize {Y}}^L$ using RSA\\
${\rm\footnotesize\overline{X}}^L=Global~pool({\rm\footnotesize{X}}^L$); ${\rm\footnotesize\tilde{X}}^L=Average~pool({\rm\footnotesize\overline{X}}^L$)\\
Compute cross entropy loss and update RSA by SGD
}\textbf{Level 2: Alternative optimization with RSA and DFA}\;
\For{all {\rm{X}} and {\rm{Y}} pairs}{
1) Compute index function $\alpha$ with traning set labels\\
2) Compute ${\left\{a_n\right\}}_{n=1}^N$ in DFA following Eq. \eqref{con:8} \\
3) Compute the reconstruction error loss in Eq. \eqref{con:7}\\
4) Update the parameters $\theta$ by SGD following Eq. \eqref{con:9}
}
\end{algorithm}

%% file: 4_Approach4.tex
\subsection{Bi-level Training and Alternative Optimization}

We note that the DFA is nonparametric and does not have a training phase. For the closed-set identification, there are no inter-set interaction, and the RSA only model is applied. The nonparametric frame-wise average pooling is applied to ${\rm \small\overline{X}}^L$, which aggregates features into a fixed-size representation ${\rm\small\tilde{X}}^L\in \mathbb{R}^{1\times 1\times D}$, followed by a softmax output layer to calculate the cross-entropy loss. For the verification or open-set identification, the RSA only model is trained by contrastive loss as in \cite{yang2017neural} using ${\rm\small\tilde{X}}^L$, while the PIFR is trained following a bi-level scheme using paired ${\rm \small\overline{X}}^L\&{\rm \small\overline{Y}}^L$.

First, we pre-train the RSA module following closed-set identification setting. In the second stage, the DFA module is added, and the dictionary learning similarity is used as the fine-tunning signal. The loss function is defined as 
\vspace{-7pt}
\begin{align} \setlength{\abovedisplayskip}{-6pt}\setlength{\belowdisplayskip}{-15pt}
\mathcal{L}(\theta,{\rm A})=\frac{\alpha}{N}{\parallel {\rm \small\overline{X}}^L - {\rm \small\overline{Y}}^L{\rm A}\parallel}{}_2^2 +\frac{\lambda}{N\times M}{\parallel{A}\parallel}_p \label{con:7}\vspace{-15pt}
\end{align}where $\theta(\Omega^l,\Psi,\Phi)$ denotes the parameters in RSA and implements as in \cite{wang2018non,zhou2017temporal}, $\alpha$ is an indicator function which takes value 1 when the probe and gallery set have the same identity and value -1 for different identity. $\lambda$ is usually set to 1 for simply. This objective explicitly encourages the feature sets from the same subject to be close while the feature sets of different subjects are far away from each other. 

\begin{figure}[t]\setlength{\abovecaptionskip}{0.cm}
\setlength{\belowcaptionskip}{-10mm}
\centering
\includegraphics[height=3.8cm]{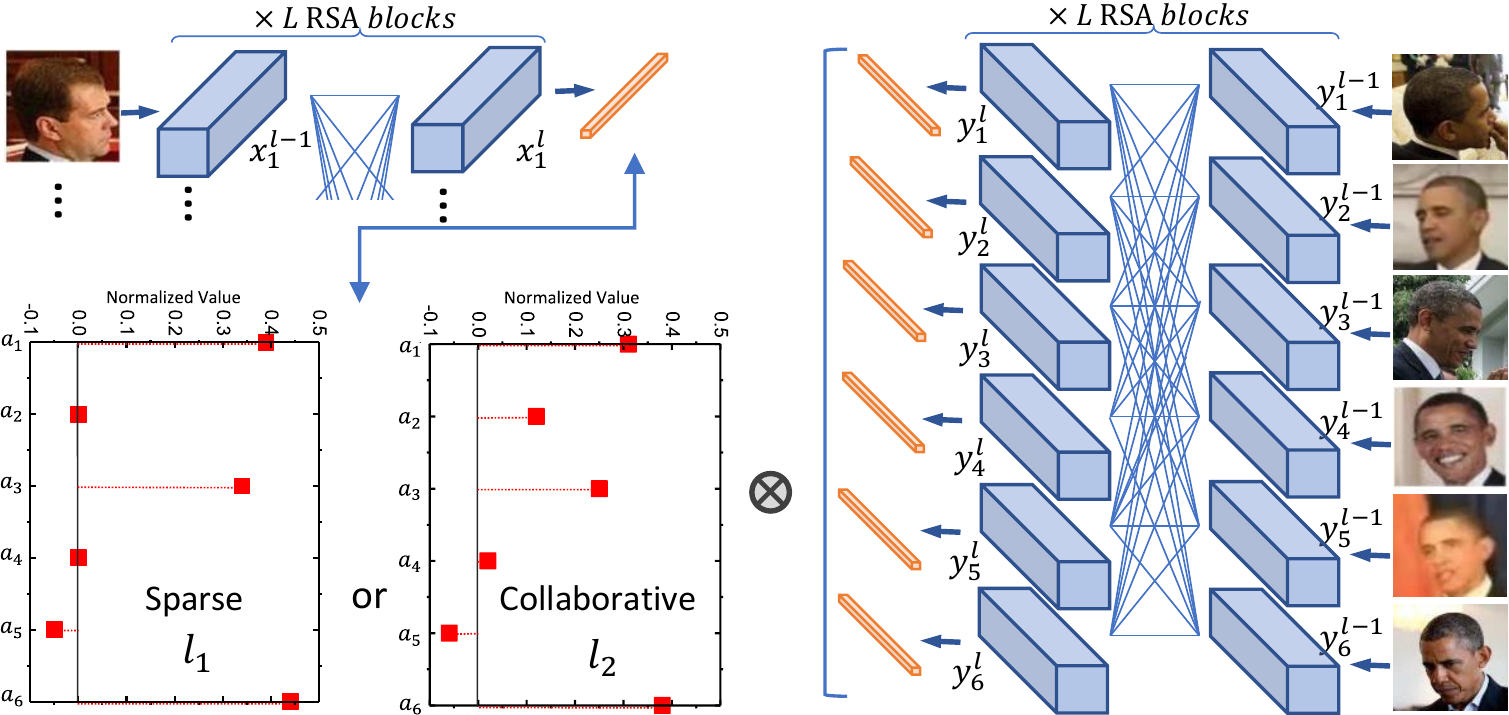}\\
\caption{Examples of DFA. The coefficients almost align the pose variation, but essentially take all possible inter-set variations into account. Following \cite{liu2018dependency}, we rotate images to frontal/right face in pre-processing stage. The value of red points $a_n$ represents the weight of each gallery feature and is a row of the DFA matrix in Fig. \ref{fig:3}. The CRC has fewer 0 values than SRC.}
\label{fig:5}
\end{figure}

To minimize the loss in Eq. \eqref{con:7}, we use the following alternative optimization algorithm to incorporate a parametric free module into deep models: 

\noindent \textbf{a) Fix (initialize) $\theta$ and minimize $\mathcal{L}$ w.r.t. A.} The task in this step is to solve the optimal sparse/collaborative coding matrix A. In practice, this is achieved by solving $a_1,\cdots a_N$ with parallel computing mode. Eq. \eqref{con:6} can be further rewritten as follows:\vspace{-7pt}\begin{align}
\mathop{}_{a_n}^{\rm min} \frac{1}{2}{a_n^T}{{\rm \small\overline{Y}}^L}^T{\rm \small\overline{Y}}^L{a_n}-{\small\overline{x}_n^T}{\rm \small\overline{Y}}^L{a_n}+\frac{\lambda}{M}{\Vert{a_n}\Vert}_p\label{con:8}\vspace{-5pt}
\end{align}~~~~~The feature-sign searching algorithm \cite{lee2007efficient,zhang2011sparse} is utilized to find the optimal ${a_n}$.

\noindent \textbf{b) Fix A and minimize $\mathcal{L}$ w.r.t. $\theta$.} One of the remarkable properties of $\mathcal{L}$ is differentiable. We update the parameters of RSA using the standard Stochastic Gradient Descent (SGD) by the following gradients of $\mathcal{L}$ w.r.t. $\rm \small\overline{X}$ and $\rm \small\overline{Y}$ \vspace{-5pt}

\begin{equation} \setlength{\belowdisplayskip}{-10pt}
\left\{  
             \begin{array}{lr}  
            \frac{\partial \mathcal{L}(\theta)}{\partial{\rm\overline{X}}}=\frac{2\alpha}{N}({{\rm \small\overline{X}}^L-{\rm \small\overline{Y}}^L{\rm A}}) & \vspace{-5pt} \\ {\tiny~} \\  \vspace{-5pt}
             \frac{\partial \mathcal{L}(\theta)}{\partial {\rm\overline{Y}}}=\frac{-2\alpha}{N}({{\rm \small\overline{X}}^L-{\rm \small\overline{Y}}^L{\rm A}}){\rm A}^T &    
             \end{array}  
\right.\label{con:9}  
\end{equation} ~~~~~The overall bi-level training is outlined in Algorithm 1. 
The similarity value learned through PIFR are self-adaptive and sensitive to spatial location and inner/inter set variations. The benefits of utilizing both of the RSA and DFA restructuring modules are demonstrated in experiments.

\noindent\textbf{Analysis.} Since IFE extracts ${\left\{x_n\right\}}_{n=1}^N$ from ${\left\{p_n\right\}}_{n=1}^N$ independently, RSA used to reconstruct the representation set is permutation-invariant and the computing of the $a_n$ in DFA is a permutation-invariant operation. Combined with the fact that the similarity measurement is permutation-invariant as well, we have the following:

\noindent \textbf{Proposition 1.} The PIFR is permutation-invariant.

Note that we can deconstruct PIFR into two operation stages to align with Eq. \eqref{con:1}. In RSA, the residual term is our $\varphi$ and sum operation, which does not act independently on each item but considers the whole set to obtain the embeddings. In the second stage, the RSA can be regarded as $\varphi$. The sum operator is inherent in $\parallel\cdot\parallel_2^2$ and $\ell_p$-norm. Then $\rho$ is simply the $sqrt$ function. Additionally, instead of a fixed function such as pooling, our inner-set relationship function is parameterized and can adapt to the task at hand.

%% file: 5_Experiments1.tex
\section{Numerical Experiments}

In this section, we first explore the meaning of the coefficients in DFA (see Fig. \ref{fig:5}). Based on the above understanding, we evaluate PIFR on several benchmarks via exhaustive ablation studies. RSA is implemented on our Titan Xp GPU. In testing stage, our gallery ${\rm\scriptsize\overline{Y}}^L$s are pre-computed and stored. The average verification time of collaborative DFA is only 93ms for IJB-A, when we use a single Xeon E5 v4 \textbf{CPU}. The detailed architectures for the individual modules are provided in the {Supplementary file}.

\begin{table*}[t]  
\tiny
\newcommand{\tabincell}[2]{\begin{tabular}{@{}#1@{}}#2\end{tabular}}
\label{tab:different_nets}
\vspace{-5mm}
\begin{center}
\begin{tabular}{|c|c|c|c|c|c|c|c|c|}
    \hline
    \multirow{2}*{Method}&\multirow{2}*{IFE Backbone}&\multirow{2}*{Dataset}& \multicolumn{3}{c|}{1:1 Verification TAR}& \multicolumn{3}{c|}{1:N Identification TPIR} \\ \cline{4-9}
    &&& FAR=0.001 &FAR=0.01 & FAR=0.1 & FPIR=0.01 & FPIR=0.1 & Rank-1 \\ \hline \hline
    
   Pose-model\cite{masi2016pose}&VGGNet&WebFace&0.652$\pm$0.037& 0.826$\pm$0.018 & - & - & - & 0.840$\pm$0.012\\ 
   
   Masi~$et~al.$\cite{masi2016we} &VGGNet&WebFace&0.725& 0.886 & - & - & - & 0.906\\ 
   
   Adaptation\cite{crosswhite2017template} &VGGNet&VGGNet&0.836$\pm$0.027 &0.939$\pm$0.013 & 0.979$\pm$0.004 & 0.774$\pm$0.049 & 0.882$\pm$0.016 & 0.928$\pm$0.010 \\
   
   DR-GAN\cite{tran2017disentangled}&CASIA-Net&WebFace& 0.539$\pm$0.043& 0.774$\pm$0.027 & - & - & - & 0.855$\pm$0.015 \\ 
   
   QAN\cite{liu2017quality} &-&-& 0.893$\pm$0.039 & 0.942$\pm$0.015 & 0.980$\pm$0.006 & - & -&0.955$\pm$0.006\\
   
   NAN\cite{yang2017neural} &GoogleNet&Indoor(3M)&0.881$\pm$0.011& 0.941$\pm$0.008 & 0.978$\pm$0.003 & 0.817$\pm$0.041 & 0.917$\pm$0.009 & 0.958$\pm$0.005\\ 
   
   DAC\cite{liu2018dependency} &GoogleNet&Indoor(3M)& {0.893}$\pm${0.010}&{0.954}$\pm${0.010}& {0.981}$\pm${0.008}& {0.855}$\pm${0.042}&{0.934}$\pm${0.009}&{0.973}$\pm${0.011}\\ \hline \hline

   {\textbf{RSA}}&GoogleNet&Indoor(3M)&{0.926}$\pm${0.011}&{0.964}$\pm${0.010}& {0.986}$\pm${0.008}& {0.882}$\pm${0.042}&{0.954}$\pm${0.009}&{0.984}$\pm${0.011}\\
   
    {\textbf{DFA($\ell_1$)}}&GoogleNet&Indoor(3M)&{0.912}$\pm${0.017}&{0.957}$\pm${0.011}& {0.983}$\pm${0.005}& {0.873}$\pm${0.045}&{0.940}$\pm${0.015}&{0.976}$\pm${0.009}\\
   
    {\textbf{DFA($\ell_2$)}}&GoogleNet&Indoor(3M)&{0.914}$\pm${0.013}&{0.958}$\pm${0.006}& {0.984}$\pm${0.003}& {0.875}$\pm${0.037}&{0.940}$\pm${0.010}&{0.978}$\pm${0.008}\\
   
    {\textbf{PIFR($\ell_1$)}}&GoogleNet&Indoor(3M)& {0.933}$\pm${0.015}&{0.966}$\pm${0.007}& {0.987}$\pm${0.005}& {0.894}$\pm${0.039}&\textbf{0.958}$\pm$\textbf{0.013}&{0.985}$\pm${0.010}\\
        
    {\textbf{PIFR($\ell_2$)}}&GoogleNet&Indoor(3M)& {0.937}$\pm${0.016}&\textbf{0.968}$\pm$\textbf{0.005}& \textbf{0.988}$\pm$\textbf{0.004}& {0.896}$\pm${0.035}&\textbf{0.958}$\pm$\textbf{0.007}&\textbf{0.988}$\pm$\textbf{0.006}\\
       
    {\textbf{PIFR($\ell_2)\times$2}}&GoogleNet&Indoor(3M)& \textbf{0.938}$\pm$\textbf{0.013}&\textbf{0.968}$\pm$\textbf{0.006}& {0.988}$\pm${0.006}& \textbf{0.897}$\pm$\textbf{0.031}&\textbf{0.958}$\pm$\textbf{0.009}&\textbf{0.988}$\pm$\textbf{0.007}\\
              
   \hline  \hline

   VGG Face2\cite{cao2018vggface2} & ResNet50&VGGFace2&$0.895\pm$0.019 & 0.950$\pm$0.005 & 0.980$\pm$0.003 & - & - &-\\ 
   
   MNet\cite{xie2018multicolumn} & ResNet50&VGGFace2& 0.920$\pm$0.013 & 0.962$\pm$0.005 & 0.989$\pm$0.002 & -& - & -\\ \hline\hline
   
   {\textbf{RSA}}& ResNet50&VGGFace2 &0.943$\pm$0.008& 0.976$\pm$0.006& 0.991$\pm$0.002 & 0.886$\pm$0.041 & 0.962$\pm$0.009 & 0.987$\pm$0.005\\ 
   
   {\textbf{PIFR($\ell_2$)}}& ResNet50&VGGFace2&\textbf{0.955}$\pm$\textbf{0.010}& \textbf{0.983}$\pm$\textbf{0.004}& 	\textbf{0.993}$\pm$\textbf{0.003}& \textbf{0.908}$\pm$\textbf{0.028}&\textbf{0.969}$\pm$\textbf{0.009}&\textbf{0.990}$\pm$\textbf{0.005}\\ \hline

\end{tabular}\label{tab:first}
\end{center}
\caption{Evaluation on the IJB-A dataset. For verification, the true accept rates (TAR) vs. false positive rates (FAR) are presented. For identification, the true positive identification rate (TPIR) vs. false positive identification rate (FPIR) and the Rank-1 accuracy are reported.}
\end{table*}

%% file: 5_Experiments2.tex
\noindent\textbf{Unconstrained Set-based face recognition}

The IARPA Janus Benchmarks (IJB)-A is a set-based face recognition dataset, where all images and videos are collected in unconstrained environments with large variants in imaging condition and viewpoints. It contains 5,397 still images and 20,412 video frames from 500 identities, and the number of samples in a set ranges from 1 to 190 with an average of 11.4 images and 4.2 videos per subject.

To take advantage of the long-tested image-based network structure and millions of available training images, the backbone of IFE is adapted from the GoogleNet \cite{szegedy2016rethinking} /ResNet50 \cite{he2016deep} and trained on 3M Indoor face images \cite{yang2017neural} or VGGFace2 \cite{cao2018vggface2} (3.31M images) respectively. We choose the same training details as \cite{liu2018dependency,xie2018multicolumn} for fair comparisons.

In the baseline methods, CNN+Mean$\ell_2$ measures the $\ell_2$ similarity of all ${\rm\footnotesize\overline{X}}^L$ and ${\rm\footnotesize\overline{Y}}^L$ pairs. The CNN+AvePool is the average pooling along each feature dimension for aggregation, followed by a linear classifier \cite{yang2017neural,liu2018dependency}.

\begin{figure}[t]
\centering
\includegraphics[height=3.4cm]{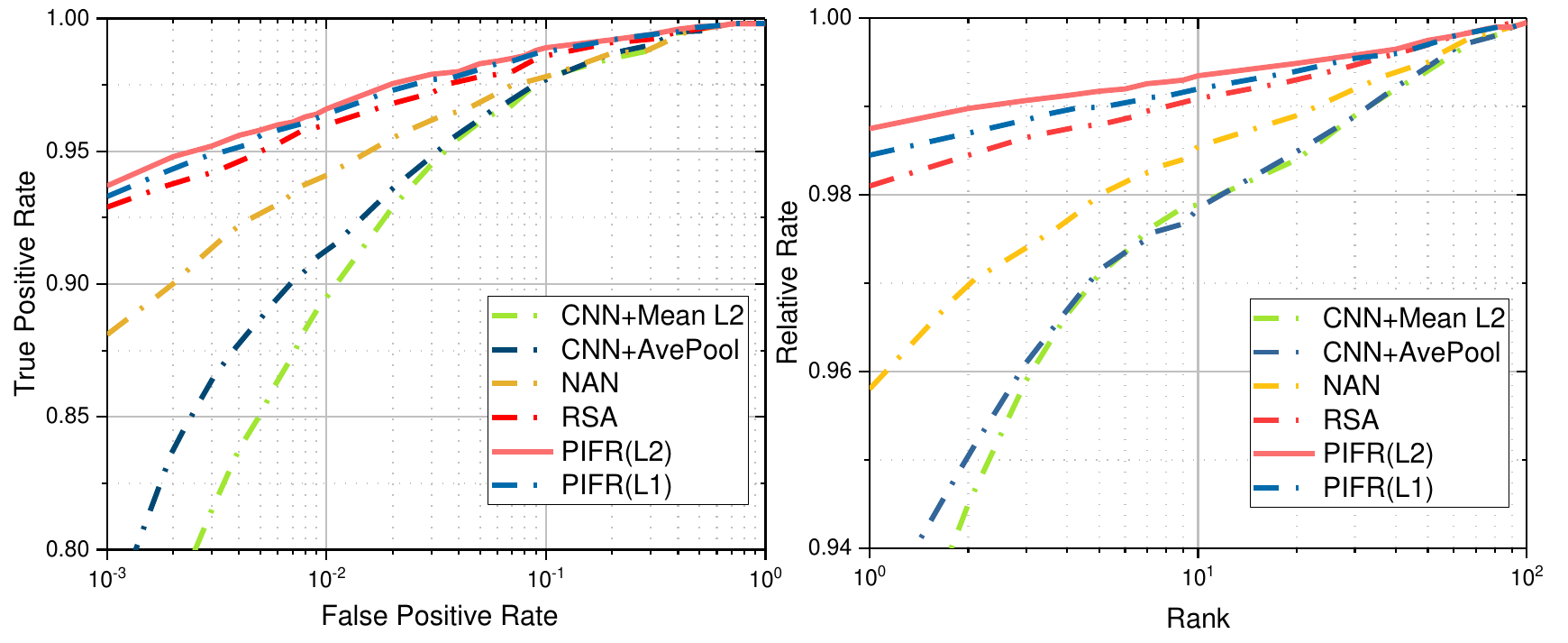}\\
\caption{Average ROC (Left) and CMC (Right) curves of the proposed method (with GoogleNet backbone) and its baselines on the IJB-A dataset over 10 splits.}
\label{fig:7}
\end{figure}

For the 1:1 $verification$ task, the performances are shown using the receiver operating characteristics (ROC) curves in Fig. \ref{fig:7}. We also report the true accept rate (TAR) $vs.$ false positive rates (FAR) in Table \textcolor[rgb]{1,0,0}{3}. 

The performance of 1:N $identification$ is characterized by the cumulative match characteristics (CMC) curves shown in Fig. \ref{fig:7}. It is an information retrieval metric, which plots identification rates corresponding to different ranks. A rank-$k$ identification rate is defined as the percentage of probe searches whose gallery match is returned within the top-$k$ matches. The true positive identification rate (TPIR) $vs.$ false positive identification rate (FPIR), as well as the rank-1 accuracy, are shown in Table \textcolor[rgb]{1,0,0}{3}.

The GoogleNet-based RSA, DFA($\ell_2$), PIFR($\ell_2$) outperforms DAC \cite{liu2018dependency} by 2.6\%/3.3\%, 2.0\%/2.1\%, \textbf{4.2\%/4.5\%} in terms of TPIR@FPIR=$10^{-2}$/TAR@FAR=$10^{-3}$ respectively. We note that the training and testing time of PIFR($\ell_2$) are 3$\times$ and 1.5$\times$ smaller than DAC \cite{liu2018dependency}. These results not only show that our model improve both the verification and identification performance significantly, but also indicates the RSA and DFA can be well combined in our alternative optimization framework and are complimentary to each other. The improvement over the previous state-of-the-art $w.r.t.$ rank-1 accuracy is also more notable than of recent works \cite{yang2017neural,liu2018dependency}. Using the ResNet50 backbone, our PIFR($\ell_2$) achieves state-of-the-art 99.0\% and outperforms \cite{xie2018multicolumn} by 3.5\% for TAR@FAR=$10^{-3}$. 

The performance of PIFR($\ell_2$) slightly exceeds PIFR($\ell_1$), indicating that the collaborative representation is better for set feature reconstructions. A possible reason is that SRC usually assigns the coefficient strictly to 0 (see Fig. \ref{fig:5}), which is similar to hard rather than soft attention \cite{xu2015show}.

The PIFR($\ell_2)\times$2 represents the symmetric dictionary learning which reconstructs $\overline{x}_n^{L}$ using $\left\{\overline{y}_1^{L},\cdots,\overline{y}_M^{L}\right\}$, and reconstructs $\overline{y}_m^{L}$ using $\left\{\overline{x}_1^{L},\cdots,\overline{x}_N^{L}\right\}$ simultaneously. However, the doubled complexity does not lead to improved performance. Although the GoogleNet and ResNet50 are designed in different structures and trained on different datasets, PIFR can be applied on top of either of them.

\begin{figure}[t]\setlength{\belowcaptionskip}{-20mm}
\centering
\includegraphics[height=3.3cm]{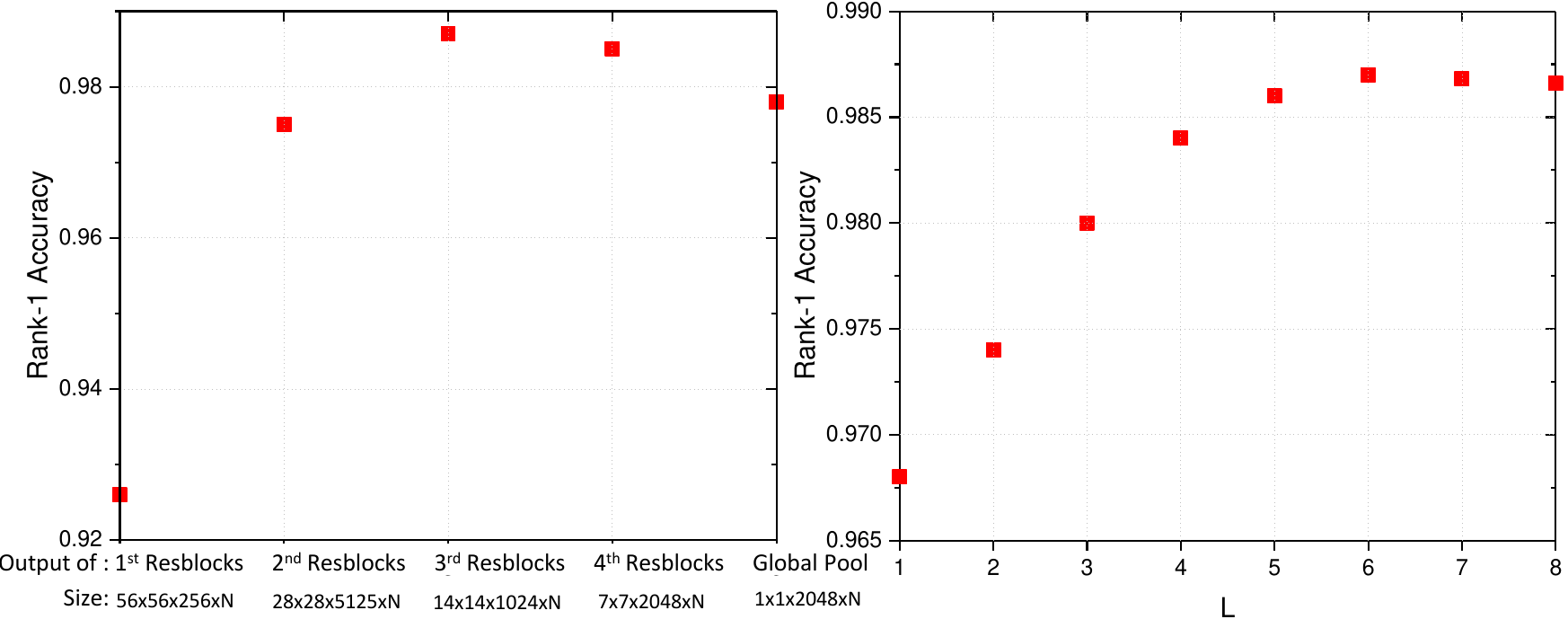}\\
\caption{On IJB-A dataset, selecting the output layer of IFE (left) and the number of RSA blocks (right) matters. The low-level features are less discriminative, while the higher-level features lose too much spatial information. The performance usually saturated with more than 5 RSA blocks.}
\label{fig:6}
\end{figure}

Then PIFR's sensitivity to the level of feature extraction and the number of RSA blocks is analyzed in Fig. \ref{fig:6}. 

%% file: 5_Experiments3.tex
\noindent\textbf{Unconstrained video-based face recognition}

The Celebrity-1000 dataset is a video-based unconstrained face identification dataset, but used as a typical set-based benchmark. It contains 2.4M frames from 159,726 videos of 1,000 subjects. We follow the protocol in \cite{yang2017neural,liu2018dependency} and report the performances in Table \ref{tab:4}$\&$\ref{tab:5} and Fig. \ref{fig:8}.

In the closed-set testing, we choose the restructured representations from the $5^{th}$ RSA blocks, and use the softmax output of our RSA only model. The RSA outperforms all of the previous methods consistently following the setting without fine-tuning the IFE module on Celebrity-1000 dataset. The video sequences are obviously redundant, and the inner-set correlation modeling does contribute to the improvements over \cite{yang2017neural}. The low quality of the several former frames also have negative impact of \cite{liu2018dependency}.

\begin{table}
\begin{center}
\tiny
\begin{tabular}{|l|c|c|c|c|c|c|}
\hline
\multirow{2}*{Method} &Temp&\multicolumn{4}{c|}{Number of Subjects($closed$)} \\\cline{3-6}
&info& 100 & 200 & 500 & 1000\\
\hline\hline
NAN(GoogleNet)\cite{yang2017neural}&no& 0.9044&0.8333&0.8227&0.7717\\\hline
DAC(GoogleNet)\cite{liu2018dependency}&no&{0.9137}&{0.8783}&{0.8523}&{0.8353}\\\hline\hline

RSA(GoogleNet)&no&\textbf{0.9385}&\textbf{0.9047}&\textbf{0.8859}&\textbf{0.8572}\\\hline\hline

RSA(ResNet50)&no&\textbf{0.9458}&\textbf{0.9140}&\textbf{0.9943}&\textbf{0.8671}\\\hline
\end{tabular}
\end{center}
\caption{Closed-set identification performance (rank-1 accuracies), on the Celebrity-1000 dataset. This setting does not have inter-set comparison, and only RSA is necessary. }\label{tab:4}
\end{table}

\begin{table}
\tiny
\begin{center}
\begin{tabular}{|l|c|c|c|c|c|c|}
\hline
\multirow{2}*{Method}&Temp&\multicolumn{4}{c|}{Number of Subjects($open$)} \\\cline{3-6}
&info& 100 & 200 & 500 & 800\\
\hline\hline

CNN+Mean $\ell_2$(GoogleNet)&no&0.8488&0.7988&0.7676&0.7067\\\hline
AvePool(GoogleNet)&no& 0.8411&0.7909&0.7840&0.7512\\\hline
NAN(GoogleNet)\cite{yang2017neural}&no& 0.8876&0.8521&0.8274&0.7987\\\hline
DAC(GoogleNet)\cite{liu2018dependency}&no&{0.9004}&{0.8715}&{0.8428}&{0.8264}\\\hline\hline

RSA(GoogleNet)non-local&no&{0.8987}&{0.8680}&{0.8376}&{0.8251}\\\hline

RSA(GoogleNet)w/o$\Delta$&no&{0.9104}&{0.8812}&{0.8545}&{0.8420}\\\hline

RSA(GoogleNet)-$x_{n\_j}^{l-1}$&no&{0.9097}&{0.8791}&{0.8603}&{0.8388}\\\hline

RSA(GoogleNet)-$(\Psi\Phi)^L$&no&{0.9297}&{0.8990}&{0.8764}&{0.8556}\\\hline

RSA(GoogleNet)&no&{0.9297}&{0.8991}&{0.8763}&{0.8558}\\\hline

DFA($\ell_2$)(GoogleNet)&no&{0.9105}&{0.8807}&{0.8562}&{0.8383}\\\hline

PIFR($\ell_2$)(GoogleNet)&no&\textbf{0.9355}&\textbf{0.9189}&\textbf{0.8876}&\textbf{0.8604}\\\hline\hline

PIFR($\ell_2$)(ResNet50)&no&\textbf{0.9497}&\textbf{0.9308}&\textbf{0.8922}&\textbf{0.8713}\\
\hline

\end{tabular}
\end{center}
\caption{Open-set identification performance (rank-1 accuracies) on the Celebrity-1000 dataset}\label{tab:5}
\end{table}

For the open-set protocol, we further measure the collaborative representation similarity to rank the gallery subjects. We see that our GoogleNet based PIFR outperforms the average pooling baseline and DAC by more than 9\% and 3\% at all measurements. More appealingly, the use of ResNet50 can further enhance our framework and achieve the state-of-the-art results with more than 4\% improvement over \cite{liu2018dependency}.

We did not manage to get competitive results using original non-local \cite{wang2018non} in our RSA. We note that non-local is also first introduced to set-based recognition in this paper, and can be regarded as a baseline. For ablation study, in RSAw/o$\Delta$, we excise $\Delta$ in Eq. \eqref{con:2}. ${\footnotesize x_{n\_j}^{l-1}}$ refers to using ${\footnotesize x_{n\_j}^{l-1}}$ as residue term instead of the difference in Eq. \eqref{con:2}. Their inferior performance demonstrates the effectiveness of our choices. $(\Psi\Phi)^L$, denotes using embedded Gaussian pair-wise affinity for every RSA blocks, which has similar performance but takes more than 2$\times$ training time.

%% file: 5_Experiments4.tex
\noindent\textbf{Person re-identification}

To verify the effectiveness of PIFR on full body images, we carry out additional experiments on the iLIDS-VID \cite{Wang_2014} benchmark which was created at an airport arrival hall under a multi-camera CCTV network, and contains 300 people in total. Its image sequences (rangings from 23 to 192 images) were accompanied by clothing similarities among people, lighting and viewpoint variations, cluttered background and occlusions. We follow the evaluation protocol from \cite{li2018diversity} for 10-fold cross evaluation. We sample the frames with a stride of 5, and the remaining frames, in the end, are simply discarded. Notice that since the data pre-processing, training setting and network structure vary in different state-of-the-art methods, we only list recent best-performing methods in the tables just for reference. For Re-ID, the widely used Cumulative Match Curve (CMC) is adopted.

\begin{figure}[t]
\centering
\includegraphics[height=3.35cm]{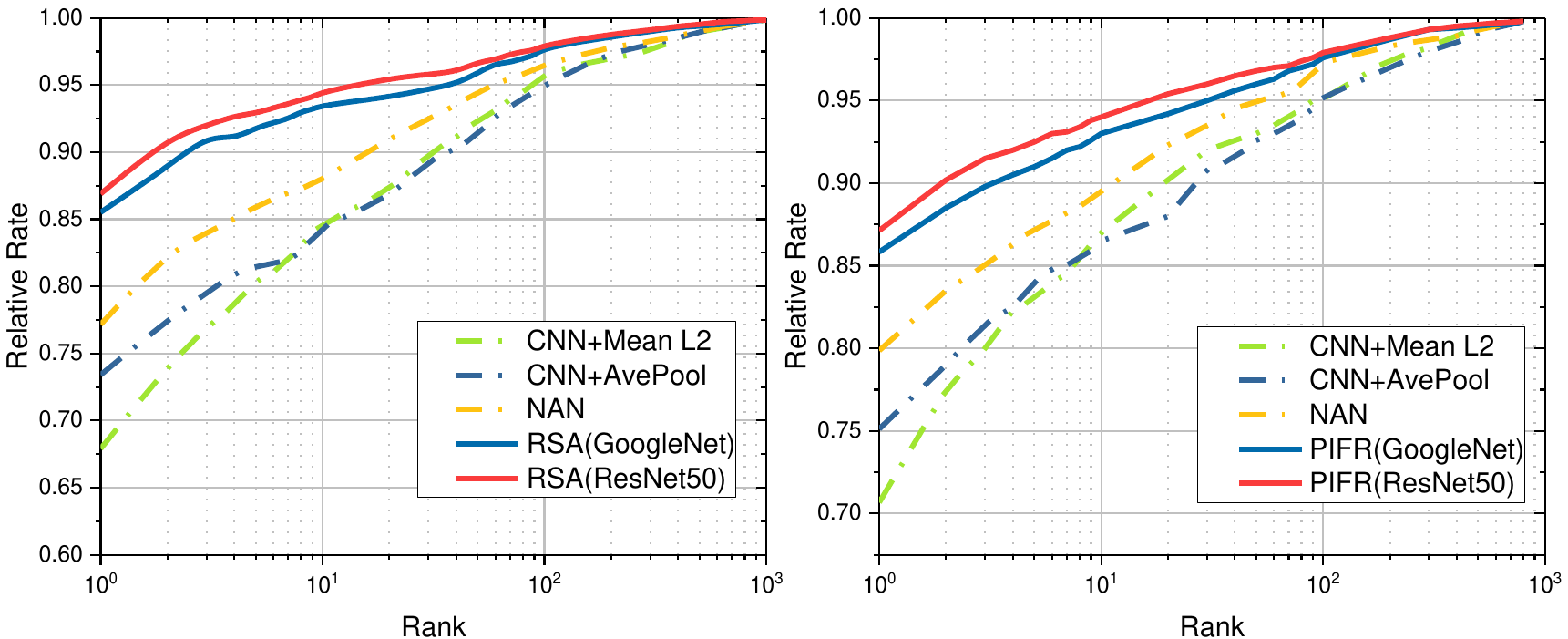}\\
\caption{CMC curves of different methods on Celebrity 1000. (left) Close-set identification on 1000 subjects, (right) Open-set identification on 800 subjects.}
\label{fig:8}
\end{figure}

We use the same IFE as in \cite{li2018diversity}, and set $L=5$. The PIFR with ResNet50 backbone is comparable to the TCP which is based on more powerful ResNet101 and extra training data. Benefiting from the more efficient inner-set complementary/diversity exploring and additional inter-set correlation module, PIFR achieves 82.5\% rank-1 accuracy dispense with the optical flow information (see Table \ref{tab:6}). It outperforms the STAN and TCP by 2.3\% and 10.8\% respectively, and 2$\times$ faster than STAN $w.r.t.$ training.

\begin{table}
\tiny
\begin{center}
\begin{tabular}{|l|c|c|c|c|c|c|}
\hline
\multirow{2}*{Method} & Temp & \multicolumn{4}{c|}{iLIDS-VID} \\\cline{3-6}
&info& CMC-1 & CMC-5 & CMC-10 & CMC-20\\
\hline\hline
TAM+SRM\cite{Zhou_2017}&yes&0.552&0.865&-&0.970\\\hline
ASTPN\cite{Xu_2017}&yes&{0.620}&{0.860}&{0.940}&{0.980}\\\hline
QAN\cite{liu2017quality}&no&{0.680}&{0.868}&{0.954}&{0.974}\\\hline
RQEN\cite{song2017regionbased}&no&{0.771}&{0.932}&{0.977}&\underline{0.994}\\\hline
STAN(ResNet50)\cite{li2018diversity}&no&\underline{0.802}&{-}&{-}&{-}\\\hline
TCP(ResNet101)\cite{liu2018transductive}&no&0.717&\textbf{0.951}&\textbf{0.983}&0.993\\\hline\hline

PIFR($\ell_2$)(ResNet50)&no& \textbf{0.825}&\underline{0.950}&\underline{0.980}&\textbf{0.997}\\
\hline
\end{tabular}
\end{center}
\caption{Experimental results of the proposed and other comparisons on iLIDS-VID re-identification dataset. The best are in bold while the second best are underlined.}
\label{tab:6}
\end{table}

%% file: 6_Conclusions.tex
\section{Conclusions}
We present a novel permutation-invariant feature restructuring framework (PIFR) for set-based representation and similarity measurement. Both the RSA and DFA module inherit the restructuring idea (in a parametric/non-parametric manner) to take into account the inner/inter-set correlations respectively. It can emphasize the discriminative images, reduces the redundancy, adaptively aligns the probe-gallery set and is scalable to open-set identification. The proposed alternative optimization can be a unified way to train the deep learning and dictionary learning jointly. The PIFR can be a general framework for the task with variable number of order-less samples. We plan to apply it to vehicle re-id, action recognition, relation reasoning etc., in the future.

\section{Acknowledgement}
The funding support from Hong Kong Government General Research Fund GRF (No.152202/14E), Natural Science Foundation of China (NSFC) (No. 61772296, 61627819, 61727818) and Shenzhen fundamental research fund (Grant No. JCYJ20170412170438636) are greatly appreciated.